\documentclass[sigconf, nonacm]{acmart}
\usepackage{multirow}
\usepackage{amsmath}
\usepackage[ruled,linesnumbered]{algorithm2e}
\usepackage{color}
\usepackage{booktabs}

\usepackage{caption}
\usepackage{graphicx}
\usepackage{float} 
\usepackage{subcaption}
\usepackage{pifont}
\usepackage{bbding}
\newcommand\vldbdoi{XX.XX/XXX.XX}
\newcommand\vldbpages{XXX-XXX}
\newcommand\vldbvolume{14}
\newcommand\vldbissue{1}
\newcommand\vldbyear{2020}
\newcommand\vldbauthors{\authors}
\newcommand\vldbtitle{\shorttitle} 
\newcommand\vldbavailabilityurl{URL_TO_YOUR_ARTIFACTS}
\newcommand\vldbpagestyle{plain} 

\begin{document}
\title{Contribution Evaluation of Heterogeneous Participants in Federated Learning via Prototypical Representations
}

\author{Qi Guo$^{*}$}
\affiliation{%
  \institution{Xi'an Jiaotong University}
  \city{Xi'an}
  \state{China}
}
\affiliation{%
  \institution{National University of Singapore}
  \city{Singapore}
  \country{Singapore}
}
\email{qiguo@u.nus.edu}

\author{Minghao Yao$^{*}$}
\author{Zhen Tian}
\author{Saiyu Qi$^{\dagger}$}
\author{Yong Qi}
\affiliation{%
  \institution{Xi'an Jiaotong University}
  \city{Xi'an}
  \state{China}
  \postcode{710129}
}
\email{minghao_yao@stu.xjtu.edu.cn}

\author{Yun Lin}
\affiliation{%
  \institution{Shanghai Jiao Tong University}
  \city{Shanghai}
  \country{China}
}
\email{lin_yun@sjtu.edu.cn}

\author{Jin Song Dong}
\affiliation{%
  \institution{National University of Singapore}
  \city{Singapore}
  \country{Singapore}
}
\email{dcsdjs@nus.edu.sg}

\thanks{$^{*}$ Equal contribution}
\thanks{$^{\dagger}$ Corresponding author. saiyu-qi@xjtu.edu.cn}

\begin{abstract}
Contribution evaluation in federated learning (FL) has become a pivotal research area due to its applicability across various domains, such as detecting low-quality datasets, enhancing model robustness, and designing incentive mechanisms. Existing contribution evaluation methods, which primarily rely on data volume, model similarity, and auxiliary test datasets, have shown success in diverse scenarios. However, their effectiveness often diminishes due to the heterogeneity of data distributions, presenting a significant challenge to their applicability. In response, this paper explores contribution evaluation in FL from an entirely new perspective of representation. In this work, we propose a new method for the contribution evaluation of heterogeneous participants in federated learning (FLCE), which introduces a novel indicator \emph{class contribution momentum} to conduct refined contribution evaluation. Our core idea is the construction and application of the class contribution momentum indicator from individual, relative, and holistic perspectives, thereby achieving an effective and efficient contribution evaluation of heterogeneous participants without relying on an auxiliary test dataset. Extensive experimental results demonstrate the superiority of our method in terms of fidelity, effectiveness, efficiency, and heterogeneity across various scenarios.
\end{abstract}

\maketitle

\pagestyle{\vldbpagestyle}
\begingroup\small\noindent\raggedright\textbf{PVLDB Reference Format:}\\
\vldbauthors. \vldbtitle. PVLDB, \vldbvolume(\vldbissue): \vldbpages, \vldbyear.\\
\href{https://doi.org/\vldbdoi}{doi:\vldbdoi}
\endgroup
\begingroup
\renewcommand\thefootnote{}\footnote{\noindent
This work is licensed under the Creative Commons BY-NC-ND 4.0 International License. Visit \url{https://creativecommons.org/licenses/by-nc-nd/4.0/} to view a copy of this license. For any use beyond those covered by this license, obtain permission by emailing \href{mailto:info@vldb.org}{info@vldb.org}. Copyright is held by the owner/author(s). Publication rights licensed to the VLDB Endowment. \\
\raggedright Proceedings of the VLDB Endowment, Vol. \vldbvolume, No. \vldbissue\ %
ISSN 2150-8097. \\
\href{https://doi.org/\vldbdoi}{doi:\vldbdoi} \\
}\addtocounter{footnote}{-1}\endgroup



\section{Introduction}
Traditional centralized deep learning, which typically relies on collecting extensive privacy-sensitive data on centralized servers, faces substantial privacy and legal challenges~\cite{GDPR2016,kairouz2021advances}. To maintain local data privacy and comply with legal regulations, federated learning (FL) emerges as a solution to enable collaborative model training across multiple participants without sharing private data~\cite{mcmahan2017communication}. FL promotes the joint collaboration of isolated data sources to achieve greater benefits and achievements~\cite{li2020review}.

When participants' data are independently and identically distributed (IID) and equal in quantity in FL, it is logical to share the same global model training outcome among participants. However, participants' data often exhibit inherent heterogeneity in practical scenarios, making it unfeasible to share the same outcome for all participants~\cite{li2021ditto,lyu2020collaborative}. Meanwhile, considering the wide applicability of contribution evaluations in detecting low-quality datasets, enhancing model robustness, and designing incentive mechanisms, etc.,~\cite{liu2022contribution,shyn2021fedccea,ding2020incentive}, it necessitates a reasonable and effective evaluation of heterogeneous participants' contributions to the FL process. Therefore, in this work, we focus on investigating the contribution evaluation of heterogeneous participants in FL, fostering the sustainable development of FL in practical applications.

Existing methods for contribution evaluation in FL typically fall into three categories: data valuation-based methods~\cite{mcmahan2017communication,li2020federated,li2021ditto}, model similarity-based methods~\cite{wang2021federated,xu2021gradient,pan2023fedmdfg}, and auxiliary test dataset-based methods~\cite{li2019fair,huang2022fairness,wang2020principled}. The rough comparison of different categories of contribution evaluation methods in FL is shown in \tablename~\ref{intro.table}. Specifically, data valuation-based methods assume that contributions are positively correlated with data valuation~\cite{mcmahan2017communication,li2021ditto}. The most straightforward approach for data valuation-based methods is to consider the volume of participant data as a standard for evaluating their contribution to FL process. However, due to the differences in data sources, collection, cleaning, and integration processes among participants in practical scenarios, the quality of data provided by each participant is inherently variable~\cite{xu2021validation}. Therefore, despite the fact that this method is efficient by directly using data valuation results, it may not be effective in contribution evaluation due to the unreliability of valuations in reality.

In model similarity-based methods, contributions are evaluated by measuring the similarity of the participant's model to the global model, such as using the $L_2$ norm distance (a smaller distance indicates a greater contribution)~\cite{wang2021federated,xu2021gradient,pan2023fedmdfg}. These methods generally evaluate contributions more effectively than data valuation-based methods. However, in practice, multiple vastly different model parameters can achieve similar local optima under various random training conditions, rendering direct comparison impractical. Additionally, large model parameters not only decrease the efficiency of contribution evaluation but also bring about the curse of dimensionality when calculating similarities~\cite{poggio2017and,guo2023fedmcsa}. This phenomenon significantly complicates the process of accurately assessing model similarity, as the vast number of parameters can distort the perception of similarity and obscure meaningful comparisons. 

For auxiliary test dataset-based methods, it is assumed that a representative auxiliary test dataset exists with the same data distribution as the data from all participants~\cite{li2019fair,huang2022fairness,wang2020principled}. The contribution of the participants can be effectively evaluated based on the accuracy of their models on this dataset. However, continuous data testing makes its efficiency low. Additionally, data is often associated with privacy concerns, acquisition difficulties, and heterogeneity. Consequently, it is highly challenging to access an ideal auxiliary test dataset that accurately represents all participants~\cite{lv2021data}.
\begin{table}[t]
\centering
\caption{Comparison of different categories of contribution evaluation methods in federated learning.}
\resizebox{1\linewidth}{!}{
\begin{tabular}{|c|c|c|c|c|}
\hline
\begin{tabular}[c]{@{}c@{}}Contribution \\ Evaluation Methods\end{tabular}         & \begin{tabular}[c]{@{}c@{}}No Auxiliary \\ Test Dataset\end{tabular} & \begin{tabular}[c]{@{}c@{}}Evaluation \\ Efficiency\end{tabular} & \begin{tabular}[c]{@{}c@{}}Evaluation \\ Effectiveness\end{tabular} & \begin{tabular}[c]{@{}c@{}}Heterogeneity \\ Treatment\end{tabular} \\ \hline

\begin{tabular}[c]{@{}c@{}}Data valuation-based\\ \cite{mcmahan2017communication,li2020federated,li2021ditto}\end{tabular}            & \CheckmarkBold                                                                    & \CheckmarkBold                                                                & \XSolidBrush                                                                   & \XSolidBrush                                                                  \\ \hline
\begin{tabular}[c]{@{}c@{}}Model similarity-based\\ \cite{wang2021federated,xu2021gradient,pan2023fedmdfg}\end{tabular}          & \CheckmarkBold                                                                    & \XSolidBrush                                                                & \CheckmarkBold                                                                   & \XSolidBrush                                                                  \\ \hline
\begin{tabular}[c]{@{}c@{}}Auxiliary test \\ dataset-based\\ \cite{li2019fair,huang2022fairness,wang2020principled}\end{tabular} & \XSolidBrush                                                                    & \XSolidBrush                                                             & \CheckmarkBold                                                                   & \CheckmarkBold                                                                  \\ \hline
\begin{tabular}[c]{@{}c@{}}Representation-based\\ {(}Our{)}\end{tabular}                 & \CheckmarkBold                                                                    & \CheckmarkBold                                                                & \CheckmarkBold                                                                   & \CheckmarkBold                                                                \\ \hline
\end{tabular}
}
\label{intro.table}
\end{table}

Apart from auxiliary test dataset-based methods, which can handle heterogeneity by testing on an ideal test dataset (noting that such a dataset is hard to obtain), the other categories have not focused sufficiently on heterogeneity. We aim to develop a new indicator that effectively and efficiently evaluates the contributions of heterogeneous participants without relying on the auxiliary test dataset. Although this goal is challenging, we have found that data representations capture the model's current learning state and data mappings. They also have the advantage of reduced dimensionality compared to the model itself~\cite{scholkopf2021toward,guo2022dual}. Therefore, we propose using data representations to evaluate the contributions of heterogeneous participants in FL. 

However, three main challenges remain. First, direct local average representations may not accurately reflect the actual contributions of heterogeneous participants due to the mutual influence of different class representations. Second, representations are dynamic and evolve towards stability, requiring consideration of their changes over rounds. Third, it is unfair not to recognize the contributions of participants not selected for training in each round.

To this end, we propose a new method for contribution evaluation of heterogeneous participants in federated learning (\textbf{FLCE}), which introduces a novel indicator \emph{class contribution momentum} to conduct refined contribution evaluation. \textbf{Our core idea is the construction and application of the class contribution momentum indicator, thereby achieving an effective and efficient contribution evaluation of heterogeneous participants without relying on an auxiliary test dataset.} Class contribution momentum consists of the \emph{class contribution mass} and \emph{class contribution velocity}, both of which derived from the average representation of the same data class. Class contribution momentum effectively mitigates interference between different data classes in heterogeneous data contribution evaluation by differentiating the impact of different data classes. It also reflects the representational mass and variation of the participant's local data in the trained model, making it an effective foundational indicator for evaluating contributions of heterogeneous participants. Furthermore, the dimensions of representations are much smaller than those of the entire model and do not grow with the number of model parameters, enabling efficient computation during the evaluation of contributions.

Specifically, FLCE evaluates contributions from three perspectives: (\romannumeral1) the individual perspective from the autonomous contribution of each participant, (\romannumeral2) the relative perspective from contribution differences across training rounds, and (\romannumeral3) the holistic perspective from the collective contribution of all participants. \textbf{From the individual perspective from the autonomous contribution of each participant}, we first compute local data representations through locally trained models of each participant and then aggregate these representations by class. The centroid of these class representations, termed the class prototype, represents the model's representational capacity for that class and signifies the mass of each class contribution. Then, these models and class prototypes are uploaded to the central server. \textbf{From the relative perspective from contribution differences across training rounds}, we consider changes in each class prototype between rounds, representing the velocity of each class contribution under model training. Based on the class contribution mass and velocity, we introduce the concept of class contribution momentum, representing the contribution of each data class. \textbf{From the holistic perspective from the collective contribution of all participants}, considering that only a subset of participants is selected for aggregation in each round, we further propose a class contribution momentum completion technique to complete missing class contribution momentums in each round. 
Meanwhile, we also consider the different importance of distinct categories. 
These three perspectives build on each other progressively, working collaboratively to effectively and efficiently evaluate the nuanced contributions of heterogeneous participants throughout the training cycle.

Extensive experimental results demonstrate FLCE's superior performance in evaluating the contributions of heterogeneous participants. FLCE exhibits high fidelity to the actual performance of the global model, effectively differentiates the contributions of heterogeneous participants, and efficiently computes contribution scores without relying on an auxiliary test dataset. 

In summary, the key contributions of our work are as follows:

(1) To the best of our knowledge, this is the first work to introduce a representation-based approach for evaluating contributions in federated learning without an auxiliary test dataset. Concurrently, we propose a novel contribution evaluation indicator \emph{Class Contribution Momentum} for contribution evaluation of federated learning. This work marks a groundbreaking shift in the paradigm of contribution evaluation research within federated learning, offering a viable and unexplored perspective.

(2) We present FLCE, a new method for evaluating contributions from heterogeneous participants in federated learning. Utilizing individual, relative, and holistic perspectives, this method enables an effective and efficient contribution evaluation of heterogeneous participants without relying on an auxiliary test dataset.

(3) Our investigation is the first to involve two critical yet previously neglected issues in federated learning contribution evaluation: the contributions of participants not selected in the current training round and the different importance of distinct categories in contribution evaluation.

(4) Our extensive experiments illustrate FLCE's superiority in evaluating contributions from heterogeneous participants in terms of performance fidelity, effectiveness, efficiency, and handling of heterogeneity.





\section{Related Works}
\subsection{Federated Learning}
Federated Learning (FL) is a new paradigm that addresses the conflict between privacy protection and knowledge acquisition by training local models across multiple decentralized participants. In this approach, instead of transferring raw data to a central server, participants train models on their own data and devices. They then upload these models to the server where they are aggregated (e.g., using FedAvg~\cite{mcmahan2017communication}, which averages the local model parameters of participants) before being redistributed. This collaborative learning method enables privacy-preserving model training without exposing sensitive data.
However, the original FL framework faces several challenges throughout the training stages~\cite{ye2023heterogeneous}. During the interaction between participants and the server, the training process may be impeded by issues such as device or network heterogeneity~\cite{xu2023asynchronous,ilhan2023scalefl,hu2023scheduling}. Additionally, the aggregation and distribution of the global model can be vulnerable to privacy breaches and poisoning attacks if malicious actors are involved~\cite{rathee2023elsa,li20233dfed,arazzi2023turning,zhou2022ppa,wu2023understanding,guo2022flmjr}. Moreover, a fundamental challenge is that the data heterogeneity among participants significantly impacts the efficiency and performance of FL~\cite{ye2023heterogeneous,li2022federated,yuan2021we}.

\subsection{Data Heterogeneity}
Data heterogeneity among participants primarily involves differences in data distribution, size, categories, and noise~\cite{ye2023heterogeneous}. Previous studies have proposed various methods to address these heterogeneity issues~\cite{xie2022federatedscope}. Tian \textit{et al.}~\cite{li2020federated} improved performance in heterogeneous environments by adding regularization, while Fang \textit{et al.}~\cite{fang2022robust} reduced the impact of noise in heterogeneous datasets by assigning weights to participants. However, these methods often struggle to fully address data heterogeneity by focusing primarily on the data itself and exploring different data types.
FedCA~\cite{zhang2023federated} was the first to merge contrastive learning with FL in an unsupervised manner, while MOON~\cite{li2021model} uses supervised contrastive learning to boost model performance. Additionally, several studies have validated the effectiveness of these representations in heterogeneous scenarios~\cite{tan2022federated,tan2022fedproto,huang2023rethinking,dai2023tackling}, mainly focusing on enhancing model performance. The improvement in performance is largely due to the reasonable allocation of local model weights, laying the groundwork for achieving greater contribution evaluations.

\subsection{Fairness}
Fairness presents a significant challenge in FL and is closely related to research on contribution evaluation. In this field, various concepts of fairness are considered, each focusing on different aspects. Some studies emphasize performance distribution fairness, which assesses consistency in performance across client devices in FL~\cite{li2019fair}. Others, such as group fairness, aim to reduce discrepancies in algorithmic decisions among diverse groups~\cite{du2021fairness,dwork2012fairness,hardt2016equality,shaham2022models,pujol2022prefair}. Additionally, some research seeks to minimize maximum loss for protected groups, thus preventing overfitting to any specific model at the expense of others~\cite{martinez2020minimax}.
However, existing fairness-oriented approaches face challenges in evaluating participant contributions in real-world scenarios. These methods struggle to accurately and efficiently evaluate participant contributions, which is crucial for attracting excellent local models for global model updates.

\subsection{Contribution Evaluation in Federated Learning}
Contribution evaluation in FL has emerged as a critical research area due to its applicability across various domains, including detecting low-quality participants, enhancing model robustness, designing incentive mechanisms, and accelerating model convergence~\cite{liu2022contribution,shyn2021fedccea,ding2020incentive,wang2022efficient}. Given the inherent challenges of data heterogeneity in FL, it is crucial to develop a reasonable and effective method for evaluating the contributions of heterogeneous participants.

Previous methods for evaluating contributions in FL can typically be grouped into three categories: data valuation-based, model similarity-based, and auxiliary test dataset-based methods. Initially, contributions can be evaluated by the volume of data from participants, with methods like FedAvg~\cite{mcmahan2017communication} and FedProx~\cite{li2020federated} assigning weights based on data size. Additionally, Ditto~\cite{li2021ditto} uses data volume to balance fairness and robustness in personalized learning. However, data volume alone may not fully reflect a participant's contribution to the global model. Thus, evaluating the similarity between local and global models becomes a viable approach. For example, FedFV~\cite{wang2021federated} mitigates potential conflicts among participants to acquire fairness; CGSV~\cite{xu2021gradient} evaluates contributions by calculating the cosine similarity between participants and the global model; FedMDFG~\cite{pan2023fedmdfg} ensures fairness by finding appropriate model update directions and step sizes. Auxiliary test datasets also play a crucial role in overcoming the limitations of data scale and model similarity evaluations due to their flexibility. For instance, q-FedAvg~\cite{li2019fair} ensures fairness by uploading cross-entropy on auxiliary test datasets; FedFa~\cite{huang2022fairness} allocates aggregate weights by uploading participant accuracy and participation frequency. Moreover, some other works depend on Game Theory to evaluate each participant's effect, they also require auxiliary test datasets and it requires a significant amount of time to calculate contribution evaluation metrics like Shapley Value~\cite{ghorbani2019data,liu2022gtg,fan2022improving,zheng2023secure}.

These methods often struggle to efficiently and effectively evaluate the contributions of heterogeneous data from participants, impacting the fairness of weight allocation during model aggregation and potentially disadvantaging some participants. Through the construction and application of the class contribution momentum indicator, our proposed method achieves an effective and efficient contribution evaluation of heterogeneous participants without relying on an auxiliary test dataset.

\section{Methodology}

\subsection{Problem Definition and Notation}

In FL, there are $n$ participants and one central server. Each participant has a local private dataset $\mathcal{D}_k=\left\{\left(x_i, y_i\right)\right\}_{i=1}^{|\mathcal{D}_k|}, k \in \{1, 2, ..., n\}$, where $x_i \in \mathbb{R}^I$ represents the $I$-dimensional feature vector of a sample, $y_i$ is the one-hot vector of the ground truth label, and $|\mathcal{D}_k|$ is the size of dataset $\mathcal{D}_k$. The goal of FL is to enable all participants to jointly train a shared global model using their individual private datasets, which can be formulated as an optimization problem:

\begin{equation}
\min _{{w}\in \mathbb{R}^d} F(w):= \sum_{k = 1}^{n} \alpha_k F_k({w}),
\label{eq:fl_goal}
\end{equation}
where $n$ is the total number of participants, ${w}\in \mathbb{R}^d$ signifies the $d$ parameters of the global model (like weights in a neural network)), $\alpha_k > 0$ with $\sum_{k}\alpha_k = 1$, $F_k(w)=\mathbb{E}_{\left(x_i, y_i\right) \sim \mathcal{D}_k}[\ell({w} ; \left(x_i, y_i\right))]$ represents the expected risk for the $k$-th participant, and $\ell({w} ; \left(x_i, y_i\right))$ is the loss function of participants.

Our study considers a classic FL scenario in which a trusted third party acts as the central server, and $n$ non-malicious participants engaged in FL are presumed to be heterogeneous. Notably, the datasets possessed by these non-malicious participants may contain some label noise and feature noise, stemming from the complexity of data collection and processing in real-world scenarios.

While the absolute contribution values of each participant may vary depending on specific tasks and settings, the normalized relative contributions among different participants exhibit universality in practical. Therefore, our contribution evaluation aims to evaluate the normalized contribution proportion of each participant relative to all participants in the entire FL process, where the sum of all participants' contribution proportions equals 1.

In FL with heterogeneous participants, our goal is to conduct an effective and efficient contribution evaluation of heterogeneous participants without relying on the auxiliary test dataset.
 
\begin{figure*}[t]
	\centering
	\includegraphics[width=1.95\columnwidth]{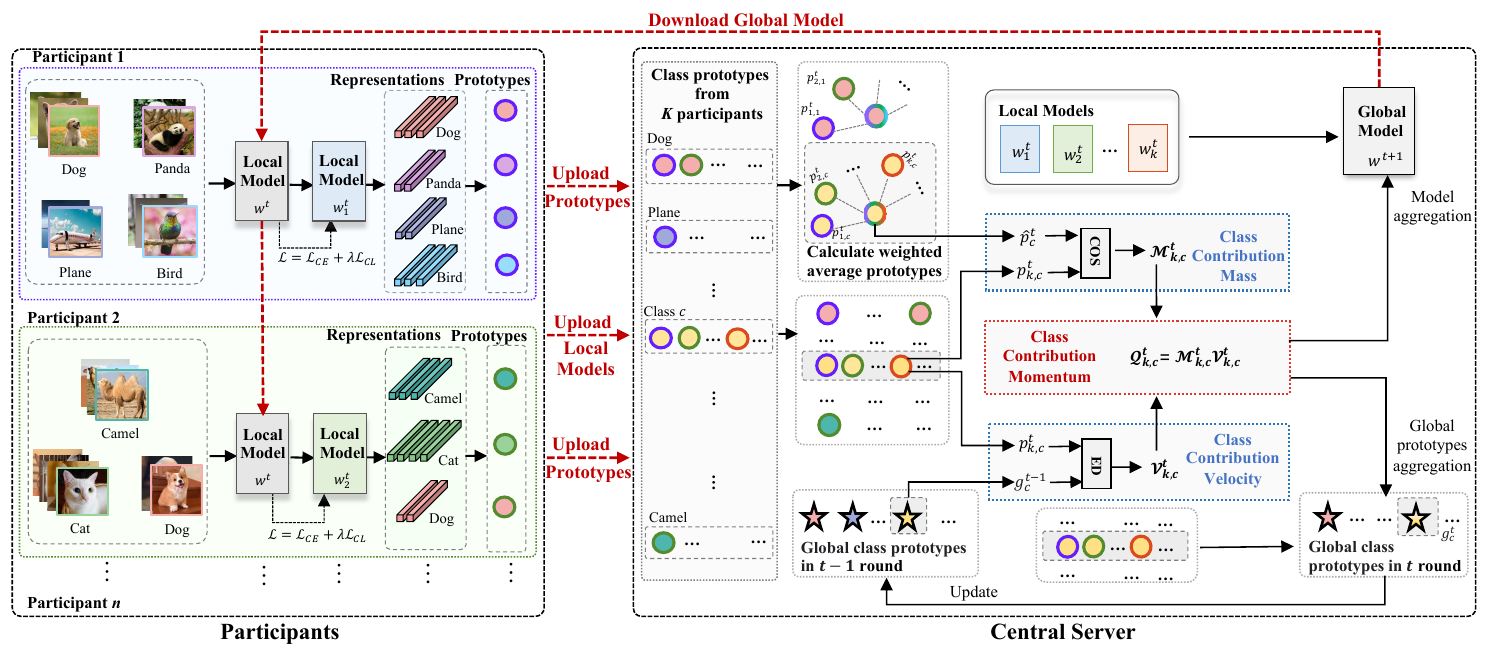}
	\caption{Framework of the proposed FLCE for contribution evaluation of heterogeneous participants in federated learning.}
\label{fig:framework}
\end{figure*}
\subsection{Overview}

We present a brief overview of our proposed method, FLCE, for the contribution evaluation of heterogeneous participants in FL, as illustrated in \figurename~\ref{fig:framework}. FLCE is a client-server architecture framework that is consistent with the standard FL framework. We adopt a tripartite perspective to conduct FLCE, encompassing the individual perspective, the relative perspective, and the overall perspective. The individual perspective focuses on the autonomous contribution of each participant. The relative perspective examines the contribution differences across training rounds. Lastly, the overall perspective considers the collective contribution of all participants. The details of FLCE are presented in the subsequent content.

\subsection{The individual perspective from the autonomous contribution of each participant}
For the contribution evaluation in FL, the most straightforward approach is to evaluate the contribution of each participant in the current training round. This reflects the individual contribution of the participants selected in each round, thus representing the individual perspective from the autonomous contribution of each participant. Directly using average representations of the participant's local data, processed through the post-training local model, may be ineffective in accurately reflecting the participant's current round contribution due to the mutual influence and interference of representations of different classes. Considering the unique data distribution of each participant, to better capture their contributions, it is important to identify both the commonality and the difference in representations among participants. The commonality lies in all participants' data sharing a common latent class-aware data distribution space. Due to the fact that each participant only has its own private data, each participant has only a subset of the complete latent class-aware distribution space. Considering that the central representation of each data class, also known as the class prototype, can be viewed as an effective representation of that class using the current model, we propose utilizing the class prototype as a reference for contribution evaluation, termed class contribution mass. From the viewpoint of class prototypes, we deconstruct the complete latent data distribution space into separate class-aware data distribution spaces. This approach effectively mitigates the interference between different class data representations within each participant. It also facilitates the collaboration of different class distributions among all participants.

Specifically, each participant, after receiving the global model from the central server, trains the model with their local data. The loss of local training for a batch of $N$ samples can be expressed as follows:
\begin{equation}
	\mathcal{L} =  \mathcal{L}_{CE} + \lambda \mathcal{L}_{CL},
    \label{eq:method.loss}
\end{equation}

\begin{equation}
\label{eq:method.lossce}
    \mathcal{L}_{CE}=-\frac{1}{N}\sum ^{N}_{i=1} {y_{i}log(\hat{y}))},
\end{equation}

\begin{equation}
	\mathcal{L}_{CL} = -\frac{1}{N}\sum_{i=1}^{N} \frac{1}{N_{y_i}} \sum_{j=1}^{N} 1_{y_i = y_j} \log \frac{e^{sim(z_i, z_j) / \tau}}{\sum_{k=1}^{N} 1_{i \neq k} e^{sim(z_i, z_k) / \tau}},
    \label{eq:method.losscl}
\end{equation}
where $\mathcal{L}_{CE}$ is the cross entropy loss, $\mathcal{L}_{CL}$ is the contrastive loss, $\lambda$ is the coefficient balancing cross entropy loss and contrastive loss, $\hat{y}$ is the probability output predicted by the model for the sample, and $z_i$ ($z_j$) denotes the representation of the input $x_i$ ($x_j$). We use an encoder to extract the representation $z$ from an input $x$. For a given $ i $-th sample, $ N_{y_i} $ is the number of samples in the batch that share the same label as $ i $-th sample. The similarity measure $ sim(z_i, z_j) $ quantifies the resemblance between the representations of $ i $-th and $ j $-th samples, and is typically computed using dot product or cosine similarity. The parameter $ \tau $ serves as a temperature scaling factor, modulating the smoothness of the distribution. The indicator function $ 1_{condition} $ yields 1 when the condition is true, and 0 otherwise. The cross entropy loss $\mathcal{L}_{CE}$ is a fundamental loss function in supervised learning. Alongside this, we introduce the contrastive loss function $\mathcal{L}_{CL}$ in supervised learning. This function aims to increase the similarity for pairs of samples with the same label and decrease it for those with different labels, which is used to improve the reliability of prototypes.

Then, the participant processes local data with the trained model to generate representations. These representations are then grouped by class to create class prototypes as follows:
\begin{equation}
\label{eq:method.z}
    \overline{z}_y = \frac{1}{N_y} \sum_{i=1}^{N} 1_{y_i = y} z_i,
\end{equation}
where the average representation for a specific class \( y \), denoted as \( \overline{z}_y \), is computed by averaging the representations of all samples correctly classified as belonging to class \( y \). Here, \( N_y \) represents the count of samples in the batch that are of class \( y \). The representation of the  \( i\)-th sample is denoted by \( z_i \), and its corresponding label is \( y_i \). The function \( 1_{y_i = y} \) acts as an indicator, equating to 1 when the label of the \( i \)-th sample matches the class \( y \), and 0 otherwise. This formulation enables the computation of the centroid of the representations for a specific class, reflecting the average location in the feature space of the samples correctly identified as belonging to that class.

Afterward, the trained local model and class prototypes are uploaded back to the central server.

In the $t$-th global training round, $K$ participants are selected for federated training. After local training is completed, the central server receives the models and prototypes uploaded by these participants. The prototype for the $c$-th class from the $k$-th participant is represented as $p^{t}_{k,c}$. The class contribution mass  ${\mathcal{M}}_{k,c}^{t}$ of the $c$-th class prototype from the $k$-th participant is calculated as follows:
\begin{equation}
\label{eq:method.prototype_m_norm}
{\mathcal{M}}^{t}_{k,c} = \frac{cos(p^{t}_{k,c}, \hat{p}^{t}_{c})}{\sum_{k = 1}^{K}cos(p^{t}_{k,c}, \hat{p}^{t}_{c})},
\end{equation}
where $\hat{p}^{t}_{c} = \sum_{k = 1}^{K}\mathcal{s}^{t}_{k,c} p^{t}_{k,c}$ represents the weighted average prototype of the $c$-th class. Here, ${\mathcal{s}}^{t}_{k,c}$ is the normalized weight of the cosine similarity of $p^{t}_{k,c}$ relative to $\frac{1}{K}\sum_{k = 1}^{K}{p} ^{t}_{k,c}$. 

Class contribution mass effectively captures and measures the unique and specific contribution of each participant in the learning process, focusing on individual class-level contributions rather than general participation. The significance of class contribution mass lies in its ability to reflect the individual and autonomous contribution of each participant within a federated learning round. This measure takes into account not only the commonality shared among all participants in terms of their latent class-aware distribution spaces but also acknowledges the unique data distributions of individual participants. Since each participant possesses only a subset of the complete latent class-aware distribution space, the class contribution mass becomes an effective metric to evaluate their specific contributions.

\subsection{The relative perspective from contribution differences across training rounds}
If there is minimal change in the class prototype relative to the previous global class prototype, it suggests a smaller contribution by the participant for that specific class in the current round. Conversely, a significant change in the class prototype indicates a larger contribution. Therefore, it is essential to consider the changes in class prototypes between consecutive rounds. This reflects the divergence of the class prototype obtained from local data training in the current round relative to the global class prototype derived from the previous round. Acknowledging the impact of these class prototype changes across rounds on contribution evaluation, we introduce the concept of class contribution velocity. On the central server, we maintain the latest global class prototypes. In the $t$-th round of global training, the most recent global prototype for $c$-th class is denoted as \(g^{t-1}_{c}\). Consequently, we can define the class contribution velocity ${\mathcal{V}}^{t}_{k,c}$ of the $c$-th class prototype from the $k$-th participant as follows:
\begin{equation}
\label{eq:method.prototype_v}
{\mathcal{V}}^{t}_{k,c} = \frac{\left \| p^{t}_{k,c} - g^{t-1}_{c} \right \|^{2}}{\sum_{k =  1}^{K} \left \| p^{t}_{k,c} - g^{t-1}_{c} \right \|^{2} } , \quad
{\mathcal{V}}^{t}_{k,c} = \frac{{\mathcal{V}}^{t}_{k,c}}{\sum_{k=1}^{K}{\mathcal{V}}^{t}_{k,c}} ,
\end{equation}
where ${\mathcal{V}}^{t}_{k,c}$ is the normalized distance between each selected participant's prototype and the global class prototype from the previous round. We then normalize ${\mathcal{V}}^{t}_{k,c}$ to obtain the class contribution velocity.

Class contribution velocity focuses on the dynamic nature of participants' contributions across successive training rounds, offering a detailed understanding of how each participant's contribution evolves during the learning process. By examining the changes in class prototypes between consecutive training rounds, class contribution velocity captures the divergence of these prototypes as they evolve. Essentially, class contribution velocity serves as a dynamic indicator, which contextualizes the participant's current contribution within the broader trajectory of the federated training process.

Class contribution mass captures the static, individual autonomous contributions in the current training round, while class contribution velocity indicates the dynamic, relative changes in contributions across successive training rounds. To enhance the evaluation of participant contributions, we introduce the concept of class contribution momentum. This concept combines class contribution mass and velocity, offering a more comprehensive view of participant engagement. Class contribution momentum is quantified as the normalized product of class contribution mass and class contribution velocity, as follows:
\begin{equation}
\label{eq:method.prototype_p}
{\mathcal{Q}}^{t}_{k,c} = {\mathcal{M}}^{t}_{k,c} {\mathcal{V}}^{t}_{k,c} , \quad
{\mathcal{Q}}^{t}_{k,c} = \frac{{\mathcal{Q}}^{t}_{k,c}}{\sum_{k=1}^{K}{\mathcal{Q}}^{t}_{k,c}} ,
\end{equation}
where ${\mathcal{Q}}^{t}_{k,c}$ denotes the class contribution momentum of participant $k$ with class $c$ in round $t$. We can use the class contribution momentum to get the global prototype $g_c^t$ of the $c$-th class at $t$-th round as follows:
\begin{equation}
\label{eq:method.aggregation_prototype}
    g_c^t=\sum_{k=1}^{K}{\mathcal{Q}}^{t}_{k,c}p^{t}_{k,c},
\end{equation}
Moreover, we can also use the class contribution momentum for model aggregation to obtain an updated global model $w^{t+1}$ as follows:
\begin{equation}
\label{eq:method.aggregation_model}
    w^{t+1}=\sum_{k=1}^{K}\frac{\sum_{c=1}^{C}\mathcal{Q}_{k,c}^{t}}{\sum_{k=1}^{K}\sum_{c=1}^{C}\mathcal{Q}_{k,c}^{t}}w_{k}^{t},
\end{equation}
where $w_{k}^{t}$ is the uploaded model by the $k$-th client at $t$-th round.

By merging class contribution mass and velocity, class contribution momentum provides a more holistic evaluation of participant contributions in FL. Class contribution momentum allows for a nuanced evaluation that captures both the immediate, static contribution of participants and their ongoing, dynamic involvement across training rounds. It not only acknowledges the immediate value brought by participants in a single round but also their evolving contribution throughout the learning process, providing a reasonable and interpretable contribution evaluation to federated training.

\subsection{The holistic perspective from the collective contribution of all participants}

With the establishment of the class contribution momentum, we can now directly calculate each participant's contribution by summing their class contribution momentums across various categories. However, there are still two issues that need to be solved.

First, only a select group of participants in FL is chosen for federated training in each round. Those not selected are excluded from the contribution evaluation. This can potentially lead to imbalances in contribution allocation due to selection strategies or randomness in the training process.

Second, the significance of contribution from each round may continually vary across different training rounds in the FL cycle. Concurrently, the importance of contributions from different data classes could also differ. Therefore, it is essential to consider both the distribution of total contributions over training rounds and the varying importance of different data classes.

In light of two issues, we must analyze contribution evaluations with a holistic perspective from the collective contribution of all participants. To tackle the first issue, we introduce a class contribution momentum completion technique. This technique, taking a global view of training across all participants, uses matrix factorization and completion to estimate the contributions of participants not selected in each training round. 

After completing all training rounds, we obtain a real contribution matrix $X$ that records the contribution ${\mathcal{Q}}^{t}_{k,c}$ for each round $t$, participant $k$, and class $c$. However, in the FL framework, only a subset of participants is chosen for each round, and those not selected contribute zero, regardless of their data mass. Ideally, if two participants have identical data, they should have the same contribution result in an ideal scenario. Yet, the selection mechanism can lead to discrepancies where non-selected clients do not contribute. To mitigate this fairness issue, we need to complement the real contribution matrix $X$ of FLCE to obtain an approximate contribution matrix $\hat{X}$ that approaches the ideal scenario\cite{candes2012exact}. The contribution matrix completion technique can be explained as follows:


\begin{equation}
\label{eq:prototype_p_complemention}
\begin{split}
 \min _{{U,V}}E = ||X - \hat{X}||^2= ||X -UV||^2  ,
\end{split}
\end{equation}
where $E$ is the error function of the distance between the real contribution matrix $X$ and the approximation matrix $\hat{X}$. To acquire approximation matrix $\hat{X}$, we perform matrix factorization $\hat{X}$ = $UV$, where ${U}$ (size $m \times k$) and ${V}$ (size $k \times n$) represent the matrices, $m$ and $n$ denote the number of rounds and participants, respectively. To expedite computation, we employ low-rank matrix factorization which $k < \min{\{m, n\}}$. The error $||X - \hat{X}||$ quantifies the difference between the real and approximated matrices.
We use gradient descent to approximately estimate their values and acquire $U$ and $V$ to compose the approximation contribution matrix without missing values. Obtaining the approximated matrix that closely resembles the real-world scenario allows our algorithm's results to improve from an unfair contribution matrix to a relatively fair result which maintains the interests of non-participating participants due to the selection mechanism.

To address how the total contribution is distributed across different rounds in the entire training cycle and the difference in the contribution importance of different data classes, we introduce two concepts: the global contribution distribution vector and the class contribution distribution vector. The global contribution distribution vector shows the spread of total contribution across different rounds. It is defined as:
\begin{equation}
\label{eq:method.prototype_a}
    A = (a_1, a_2, ..., a_{T}),
\end{equation}
where $T$ is the total number of global training rounds. When each element in the vector equals $\frac{1}{T}$ (the reciprocal of the total number of rounds), it indicates a typical scenario where contributions are evenly distributed across all rounds.

Additionally, the class contribution distribution vector indicates the significance of contributions in different classes. It is defined as:
\begin{equation}
\label{eq:method.prototype_b}
    B = (b_1, b_2, ..., b_{C}),
\end{equation}
where $C$ is the total number of categories. When each element in the vector equals $\frac{1}{C}$ (the reciprocal of the total number of categories), it suggests a common case where all categories are equally important in contribution evaluation.

Finally, we can calculate each participant's contribution in FL. The contribution of the $k$-th participant in a complete FL cycle is presented as follows:
\begin{equation}
\label{eq:contribution_each}
\mathcal{C\!E}_{k} = \sum_{t=1}^{T} a_{t}\sum_{c=1}^{C} b_{c}\mathcal{Q}_{k,c}^{t}, \quad
\mathcal{C\!E}_{k} = \frac{\mathcal{C\!E}_{k}}{\sum_{k=1}^{n}{\mathcal{C\!E}_{k}}}.
\end{equation}

As a result, we obtain the final contribution evaluation result $\left \{ \mathcal{C\!E}_{k} \right \}_{k=1}^{n}$ for all participants in FL.

\begin{algorithm}[t]
	\caption{Contribution Evaluation of Heterogeneous Participants in Federated Learning (FLCE)}\label{algorithm:flce}
	\KwIn{the global model $w$, the dataset $\mathcal{D}_k$, maximum training round $T$, and number of subset $K$.}
	\KwOut{The contribution evaluation result $\left \{ \mathcal{C\!E}_{k} \right \}_{k=1}^{n}$ and the final model $w^{T}$.}
	\textbf{Server executes:}\\
	initialize $w^{0}$\\
	\For{$t = 1$ to $T$}{
		Randomly select K participants $\left \{ i_{l} \right \}_{l=1}^{K}$ from $n$ clients\\
		\For{$k \gets i_1$ \textbf{to} $i_K$ in parallel}{
			send the global model $w^{t}$ to the participant\\
			$w^{t}_{k}$ and $\left \{ p^{t}_{k,c} \right \}_{c=1}^{C}$ $\gets$ \textbf{LocalTraining}$(t, k, w^{t})$\\
		}
		compute $\mathcal{Q}^{t}_{k,c}$ by $g^{t-1}_{c}$ and $p^{t}_{k,c}, k = i_1 \ to \ i_K, c = 1 \ to \ C$ \\
        $g^t_c$ $\gets$ Perform prototype updates by Eq.\ref{eq:method.aggregation_prototype} \\
        $w^{t+1}$ $\gets$ Perform model aggregation$(w^{t}_{k}, k = i_1 to i_K)$ by Eq.\ref{eq:method.aggregation_model}\\
	}
    compute the approximate contribution matrix $\hat{X}$ by Eq.\ref{eq:prototype_p_complemention}\\ 
    compute $\mathcal{C\!E}_{k}$ for each participant by Eq.\ref{eq:contribution_each}.\\
	Return the contribution evaluation result $\left \{ \mathcal{C\!E}_{k} \right \}_{k=1}^{n}$ and the final model $w^{T}$.\\
	\textbf{LocalTraining:}$(t, k, w^{t})$\textbf{:}\\
    \For{each batch}{
        compute $\mathcal{L} =  \mathcal{L}_{CE} + \lambda \mathcal{L}_{CL}$ by Eq.\ref{eq:method.lossce} and Eq.\ref{eq:method.losscl}\\
        $w^{t} \gets w^{t} - \eta \nabla \mathcal{L}$\\

    }
	$w^{t}_{k} \gets w^{t}$\\
	generate prototypes $\left \{ p^{t}_{k,c} \right \}_{c=1}^{C}$ by $w^{t}_{k}$\\
	Return $w^{t}_{k}$ and $\left \{ p^{t}_{k,c} \right \}_{c=1}^{C}$.
\end{algorithm}

FLCE adopts a tripartite perspective, encompassing individual, relative, and overall contributions. This comprehensive approach ensures that each participant’s contribution is evaluated from different dimensions, providing a more complete and nuanced understanding of their role in the FL process. 
The complete description of FLCE is presented in Algorithm~\ref{algorithm:flce}.

\section{Experiments}

\subsection{Experimental Setup}
\subsubsection{Datasets and Network Architecture}

We evaluated the performance of FL methods using three real-world datasets: CIFAR-10~\cite{krizhevsky2009learning}, CIFAR-100~\cite{krizhevsky2009learning}, and EuroSAT~\cite{helber2019eurosat}.

\textbf{CIFAR-10}~\cite{krizhevsky2009learning}:
This public dataset for image classification consists of 60,000 32x32 color images distributed across 10 categories. Each category has 6,000 images, with the dataset split into 50,000 training images and 10,000 testing images.

\textbf{CIFAR-100}~\cite{krizhevsky2009learning}:
This dataset is designed for image classification and covers a wide range of objects and scenes. It includes 60,000 32x32 color images distributed across 100 categories. Each category contains 500 training images and 100 test images.

\textbf{EuroSAT}~\cite{helber2019eurosat}:
This dataset for Earth observation and remote sensing image classification comprises 27,000 64x64 color satellite images from various regions in Europe. The dataset has 10 classes. Each class has 2,160 training images and 530 testing images.

In the experiment, we employ ResNet20 as the default network architecture, which includes 20 convolutional layers and is part of the residual network family~\cite{he2016deep}. 

\subsubsection{Baselines} 
We categorize nine baselines into three groups: (\romannumeral1) Data valuation-based methods: FedAvg~\cite{mcmahan2017communication}, FedProx~\cite{li2020federated}, and Ditto~\cite{li2021ditto};
(\romannumeral2) Model similarity-based methods: FedFV~\cite{wang2021federated}, CGSV~\cite{xu2021gradient} and MOON~\cite{li2021model}; (\romannumeral3) Auxiliary test dataset-based: q-FedAvg~\cite{li2019fair}, FedFa~\cite{huang2022fairness}, and FedSV~\cite{wang2020principled}.
The details of the nine baselines are presented as follows.




\textbf{FedAvg}~\cite{mcmahan2017communication}:
A foundational approach in federated learning that aggregates local model updates using a simple average, aiming to achieve a global model without sharing raw data.

\textbf{FedProx}~\cite{li2020federated}:
It enhances FedAvg by introducing a proximal term to mitigate system heterogeneity. This term penalizes the difference between local model updates and the global model, allowing for more effective learning in non-IID data environments.

\textbf{Ditto}~\cite{li2021ditto}:
It is a personalized FL framework that can trade off between the local model and the global model. Ditto can inherently provide fair contribution evaluations and robustness.


\begin{table*}[t]
\centering
\caption{Accuracy and F1 score on CIFAR-10, CIFAR-100, and EuroSAT datasets under the IID and Non-IID settings (\%). Note that the best results are marked in bold.}
\resizebox{0.98\linewidth}{!}{
\begin{tabular}{lcccccccccccc}
\hline
 & \multicolumn{4}{c}{CIFAR-10} & \multicolumn{4}{c}{CIFAR-100} & \multicolumn{4}{c}{EuroSAT}  \\
         & \multicolumn{2}{c}{IID} & \multicolumn{2}{c}{Non-IID}  & \multicolumn{2}{c}{IID}              & \multicolumn{2}{c}{Non-IID}  & \multicolumn{2}{c}{IID}  
         & \multicolumn{2}{c}{Non-IID} \\
         & Acc & F1 score & Acc & F1 score & Acc & F1 score & Acc  & F1 score  & Acc  & F1 score & Acc & F1 score \\
\hline
\multicolumn{1}{c|}{FedAvg}  & 87.39 & \multicolumn{1}{c|}{87.27} & 83   & \multicolumn{1}{c|}{82.81}                                       & 59.43 & \multicolumn{1}{c|}{57.68} & 56.18  & \multicolumn{1}{c|}{54.27}                                       & 97.45 & \multicolumn{1}{c|}{97.34} & 96.51  & 96.36\\
\multicolumn{1}{c|}{FedProx} & 87.19 & \multicolumn{1}{c|}{87.07} & 83.28  & \multicolumn{1}{c|}{83.05}                                       & 57.02 & \multicolumn{1}{c|}{56.05}  & 55.84  & \multicolumn{1}{c|}{53.91}                                       & 97.26 & \multicolumn{1}{c|}{97.12} & 96.39  & 96.25\\
\multicolumn{1}{c|}{Ditto}   & 84.58 & \multicolumn{1}{c|}{84.41} & 80.89  & \multicolumn{1}{c|}{80.64}                                       & 52.76 & \multicolumn{1}{c|}{50.57} & 51.15  & \multicolumn{1}{c|}{48.46}                                       & 96.6 & \multicolumn{1}{c|}{96.47}  & 95.14  & 94.98\\
\multicolumn{1}{c|}{FedFV}   & 87.17 & \multicolumn{1}{c|}{87.05} & 83.45  & \multicolumn{1}{c|}{0.8331}                                       & 57.99  & \multicolumn{1}{c|}{56.33} & 55.78  & \multicolumn{1}{c|}{53.66}                                       & 97.36 & \multicolumn{1}{c|}{97.24} & 96.56  & 96.45\\
\multicolumn{1}{c|}{CGSV}    & 87.5 & \multicolumn{1}{c|}{87.33} & 83.63  & \multicolumn{1}{c|}{83.45}                                       & 58.18 & \multicolumn{1}{c|}{56.29} & 56.86  & \multicolumn{1}{c|}{54.81}                                       & 97.2 & \multicolumn{1}{c|}{97.08} & 96.64  & 96.53\\
\multicolumn{1}{c|}{MOON}    & 87.6 & \multicolumn{1}{c|}{87.46} & 83.3  & \multicolumn{1}{c|}{82.99}                                        & 58.52 & \multicolumn{1}{c|}{56.66} & 56.97   & \multicolumn{1}{c|}{54.96}                                        & 97.17 & \multicolumn{1}{c|}{97.07} & 96.3  & 96.19\\
\multicolumn{1}{c|}{qFedAvg} & 87.35 & \multicolumn{1}{c|}{87.21} & 83.69  & \multicolumn{1}{c|}{83.53}                                       & 58.67 & \multicolumn{1}{c|}{56.76} & 57.03   & \multicolumn{1}{c|}{55.09}                                       & 96.15 & \multicolumn{1}{c|}{95.97} & 92.2   & 91.85\\
\multicolumn{1}{c|}{FedFa}   & 87.7 & \multicolumn{1}{c|}{87.54} & 83.02  & \multicolumn{1}{c|}{82.76}                                        & 58.82 & \multicolumn{1}{c|}{56.88}  & 56.69  & \multicolumn{1}{c|}{54.53}                                       & 97.41 & \multicolumn{1}{c|}{97.31} & 96.18  & 96.07\\
\multicolumn{1}{c|}{FedSV}   & 87.63 & \multicolumn{1}{c|}{87.54} & 83.4  & \multicolumn{1}{c|}{83.17} 
                             & 58.96  & \multicolumn{1}{c|}{57.02}  & 56.58  & \multicolumn{1}{c|}{54.5} 
                             & \textbf{97.68} & \multicolumn{1}{c|}{97.3} & 96.62 & 96.5 \\
\multicolumn{1}{c|}{FLCE}    & \textbf{89.11}  & \multicolumn{1}{c|}{\textbf{88.99}}  & \textbf{85.06}                       & \multicolumn{1}{c|}{\textbf{84.89}}                & \textbf{61.39}               & \multicolumn{1}{c|}{\textbf{59.67}}                 & \textbf{58.85}                & \multicolumn{1}{c|}{\textbf{57.1}}                & 97.66                                  & \multicolumn{1}{c|}{\textbf{97.57}}                                  & \textbf{96.79}               & \textbf{96.66} \\
\hline
\end{tabular}
}
\label{tab:normal}
\end{table*}

\textbf{FedFV}~\cite{wang2021federated}:
This method is designed to address fairness in FL. It aims to reduce potential conflicts between clients before averaging gradients. The algorithm initially utilizes cosine similarity to detect gradient conflicts and then iteratively eliminates such conflicts by modifying the direction and magnitude of the gradients.

\textbf{CGSV}~\cite{xu2021gradient}:
The approach utilizes the cosine similarity of local and global models to evaluate the contribution of participants. 

\textbf{MOON}~\cite{pan2023fedmdfg}:
This algorithm uses the similarity in model representations to enhance the local training of individual participants.


\textbf{qFedAvg}~\cite{li2019fair}:
q-FedAvg introduces a parameter 'q' to control the contributions of local models in the aggregation process. It offers a flexible approach in FL, allowing adjustments to the aggregation mechanism based on local model performance.

\textbf{FedFa}~\cite{huang2022fairness}:
It introduces a dual-momentum gradient optimization scheme, which accelerates the model's convergence speed. The proposed algorithm combines training accuracy and training frequency information to measure the weights, aiding clients in participating in server aggregation with fairer weights.

\textbf{FedSV}~\cite{wang2020principled}:
We use the canonical Shapley value to calculate the contribution of participants. Due to the computational complexity, we employ Monte-Carlo estimation of Shapley Value, which is conducted by randomly sampling participant permutations and eliminating unnecessary sub-model utility evaluations.

\subsubsection{Metrics}
\label{title.exp.setup.metrics}
In our experiments, we evaluate performance using three primary metrics: Accuracy, F1 Score, and Kullback-Leibler (KL) Divergence. 
KL Divergence is a statistical measure quantifying the dissimilarity between two probability distributions. Some previous papers have utilized metrics such as Cosine Distance ~\cite{liu2022gtg} or Euclidean distance~\cite{liu2022contribution} to assess the difference between contributions and evaluation criteria. In our study, considering the normalized relative contribution of individual participants and the overall contribution evaluation of all participants, we utilize KL Divergence to assess the effectiveness of various contribution evaluation methods in FL. Specifically, we compare the distribution of data quality against the distribution of contribution evaluation results obtained from different methods. The KL Divergence between two distributions P and Q is defined as $KL(P\Vert Q)$. $Q$ represents the distribution of data quality, with each element denoting the normalized data quality of an individual participant (i.e., the proportion of a participant's data volume relative to the total data volume across all participants). $P$ represents the distribution of contribution evaluation results, where each element is the normalized contribution proportion as determined by the evaluation method.
A smaller KL Divergence value indicates greater similarity between the two distributions, suggesting superior effectiveness of the contribution evaluation method. 
However, few FL algorithms are specifically aimed at evaluating contributions. If the baseline can directly calculate the contribution (e.g., FedSV) or aggregation weight (e.g., FedFa), we use the corresponding calculation results. Otherwise, we compute the similarity from the local model of participants to the global model and normalize it as the contribution value for the current round.



\subsubsection{Federated Learning Setting and Details}
In our experiments, we set the total number of clients at 50, with 10 clients selected per round. The training was conducted for 1000 rounds. The default Dirichlet coefficient $\delta$=0.5 for the Non-IID scenario. The accuracy and F1 score are the average test performance of the global model over the last hundred rounds.
We used a batch size of 64, a learning rate of 0.01, and a prototype size of 64.
The experiments were conducted on a server with Ubuntu 20.04.3 LTS, Intel(R) Xeon(R) Gold 6226R 2.90GHz CPU, and NVIDIA A100 Tensor Core GPU with 80G RAM. 
\subsection{Fidelity}
\label{title.exp.Fidelity}
Previous contribution evaluation methods in FL often overlooked the impact on the global model's performance, focusing mainly on objectives like contribution evaluation or fairness. For effective contribution evaluation in FL, ensuring the proposed method doesn't harm the global model's performance is crucial. Our primary focus is on our method's performance fidelity. To demonstrate this, we compared our method against nine baseline algorithms, showcasing the fidelity results in Table~\ref{tab:normal}.

In the context of CIFAR-10 and CIFAR-100 datasets, FLCE demonstrates superior performance in both IID and Non-IID settings. Specifically, for CIFAR-10, FLCE achieves the highest accuracy and F1 score, marking 89.11\% and 88.99\% for IID settings and 85.06\% and 84.89\% for Non-IID settings, respectively. This outperforms the second-best model, FedSV, which achieves a notable but lower accuracy and F1 score in the Non-IID setting for CIFAR-10. The performance trend is consistent in the CIFAR-100 dataset.
Turning our attention to the EuroSAT dataset, FLCE continues to exhibit exemplary performance, especially in the Non-IID setting where it achieves the highest F1 score of 96.66\% and a top accuracy of 96.79\%. Notably, while FedSV shows a marginally better accuracy in the IID setting with 97.68\%, FLCE's performance remains competitive, with an accuracy of 97.66\% and the highest F1 score of 97.57\%, illustrating its consistent effectiveness across different types of datasets. The higher performance of FedSV is mainly due to the extensive computation and verification based on the auxiliary test dataset.

The experimental results affirm the excellence and generalizability of FLCE in terms of performance fidelity. FLCE's approach demonstrates that it is possible to achieve a balance between accurately evaluating contributions and enhancing the overall model performance. The reason is that FLCE enables better contribution evaluation (see in \ref{title.exp.Effectiveness}) and thus better weight distribution of client models, which leads to consistently superior model performance.
\begin{figure}[t]
	\centering
	\includegraphics[width=0.9\linewidth]{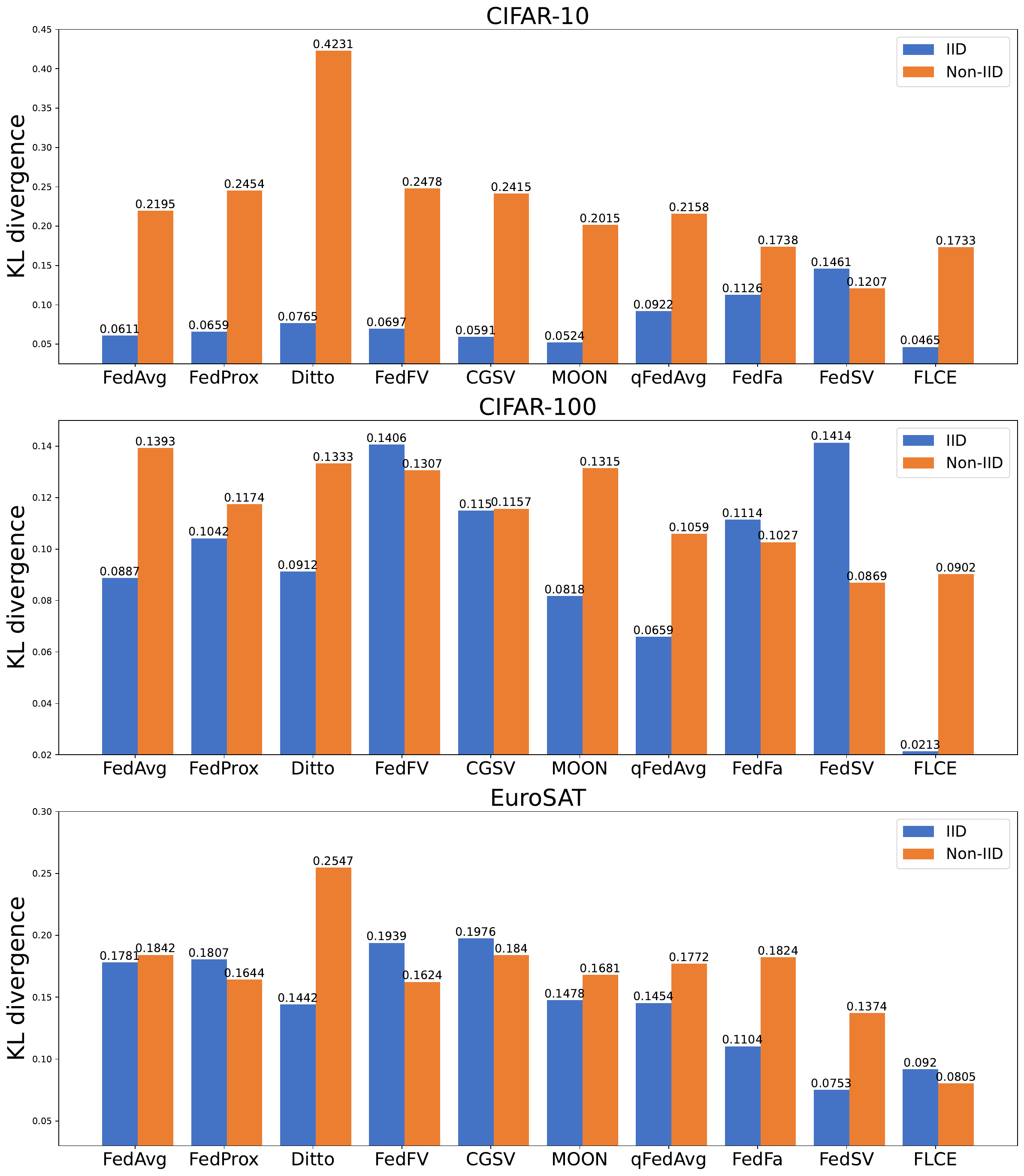}
	\caption{Effectiveness of various methods in contribution evaluation of heterogeneous participants in FL.}
	\label{fig.main KL divergence}
\end{figure}

\subsection{Effectiveness}
\label{title.exp.Effectiveness}
The purpose of this experiment is to assess the effectiveness of FLCE in the contribution evaluation of heterogeneous participants. This assessment is crucial to ascertain the method's ability to provide fair and accurate contribution evaluations across diverse scenarios. The metric is to calculate the Kullback-Leibler (KL) divergence between the contribution distribution calculated by the algorithm and the actual contribution distribution. If the divergence is smaller, it means that the two distributions are closer, then the contribution evaluation effect is better.

The effectiveness results, depicted in the \figurename~\ref{fig.main KL divergence}, show the KL divergence scores for the FLCE method compared to nine baselines across CIFAR-10, CIFAR-100, and EuroSAT datasets in both IID and Non-IID settings. With the exception of FedSV's method, FLCE consistently achieves the lowest KL divergence scores, indicating closer alignment between participants' actual contributions and their evaluations. The Shapley value-based methods have remained at the forefront of performance and contribution evaluation effectiveness due to their utilization of additional huge amounts of computing and verification resources. To the best of our knowledge, this is the first work that a non-Shapley-based approach in contribution evaluation has surpassed Shapley value-based methods in certain scenarios. The FLCE method demonstrates superior effectiveness in contribution evaluation, evidenced by its consistently lower KL divergence scores across all datasets and settings. This suggests a more accurate and fair evaluation of participants' contributions. This success is largely due to its innovative use of the class momentum contribution, which allows for a more nuanced evaluation of contributions that consider the quality and category of data each participant provides. Unlike traditional methods that might oversimplify the contribution evaluation process, FLCE's approach ensures a more equitable and comprehensive evaluation, leading to enhanced model performance (as discussed in \ref{title.exp.Fidelity}) and contribution evaluation effectiveness.

\subsection{Ablation Studies}
\label{title.exp.Ablation}

\begin{table}[t]
\centering
\caption{KL divergence on different variants of FLCE on various datasets with the IID and Non-IID settings. ``$\Delta$" refers to the change value of the variants compared with FLCE.}
\resizebox{0.98\linewidth}{!}{
\begin{tabular}{c|cc|cc|cc}
\hline
\multirow{2}{*}{Dataset} & \multicolumn{2}{c|}{CIFAR-10}      & \multicolumn{2}{c|}{CIFAR-100}     & \multicolumn{2}{c}{EuroSAT}               \\ 
                         & IID             & Non-IID          & IID             & Non-IID          & IID             & Non-IID                    \\ \hline
FLCE                     & \textbf{0.0465} & \textbf{0.1733} & \textbf{0.0213} & \textbf{0.0902} & \textbf{0.092} & \textbf{0.0805}  \\ \hline
$\text{FLCE}^{-\mathcal{M}}$                   & 0.0492          & 0.1796          & 0.0452          &0.1045           & 0.0951           & 0.128                  \\
$\Delta$                 & $\uparrow$0.0027         & $\uparrow$0.0063         & $\uparrow$0.0239         & $\uparrow$0.0143          & $\uparrow$0.0031         & $\uparrow$0.0475                 \\ \hline
$\text{FLCE}^{-\mathcal{V}}$                   & 0.055          & 0.1833          & 0.049          & 0.104          &0.1046           & 0.1192                  \\
$\Delta$                 & $\uparrow$0.0085         & $\uparrow$0.01         & $\uparrow$0.0277          & $\uparrow$0.0138         & $\uparrow$0.0126         & $\uparrow$0.0387                  \\ \hline
$\text{FLCE}^{-\mathcal{Q}}$                   & 0.065          & 0.193            & 0.0554          & 0.1          & 0.1054          & 0.116                   \\
$\Delta$                 & $\uparrow$0.0187         & $\uparrow$0.0197         & $\uparrow$0.0341         & $\uparrow$0.0098         & $\uparrow$0.0134         & $\uparrow$0.0355              \\ \hline
$\text{FLCE}^{-\mathcal{CL}}$                 & 0.0482          & 0.193          & 0.0656          & 0.134          & 0.093          & 0.0839                   \\
$\Delta$                 & $\uparrow$0.0017         & $\uparrow$0.0197         & $\uparrow$0.0443         & $\uparrow$0.0438         & $\uparrow$0.001         & $\uparrow$0.0034                \\ \hline
$\text{FLCE}^{-\mathcal{CMC}}$          & 0.0834            & 0.1869          & 0.0751          & 0.139          & 0.1623           & 0.1171                    \\
$\Delta$                 & $\uparrow$0.0369         & $\uparrow$0.0136         & $\uparrow$0.0538         & $\uparrow$0.0488          & $\uparrow$0.0703         & $\uparrow$0.0366                \\ \hline

\end{tabular}
}

\label{tab:ablation}
\end{table}

To better understand FLCE, we conducted ablation studies to evaluate the impact of its key components. Each experiment was set up identically, except for the variable of interest being tested. We constructed five variants of FLCE as follows:

1) $\text{FLCE}^{-\mathcal{M}}$: This variant removes the class contribution mass component on the server.

2) $\text{FLCE}^{-\mathcal{V}}$: This variant removes the class contribution velocity component on the server.

3) $\text{FLCE}^{-\mathcal{Q}}$: This variant removes the entire class contribution momentum component on the server.

4) $\text{FLCE}^{-\mathcal{CL}}$: This variant removes the contrastive loss component from the local training in the clients.

5) $\text{FLCE}^{-\mathcal{CMC}}$: This variant removes the contribution matrix completion component.

As shown in \tablename~\ref{tab:ablation}, we compared the KL divergence of FLCE and its five variants in the IID and Non-IID settings across CIFAR-10, CIFAR-100, and EuroSAT datasets. Notably, the removal of any component leads to an increase in KL divergence scores, signifying a drop in the ability of contribution evaluation. The $\Delta$ values indicate the relative degradation in contribution evaluation compared to the full FLCE method. The results demonstrate the individual importance of each FLCE component in reducing KL divergence, thus improving the fairness and accuracy of participant contribution evaluations. Particularly, the removal of class contribution momentum and the class contribution completion show significant increases in KL divergence, highlighting their critical roles in the FLCE approach. 

The ablation experiments not only illustrate the effectiveness of class contribution momentum, but also the individual contributions of class contribution mass and class contribution velocity. Furthermore, the findings affirm the enhancement brought by integrating contrastive loss and contribution matrix completion techniques in the class contribution momentum-based evaluation method.

\begin{figure}[t]
	\centering
	\includegraphics[width=0.9\linewidth]{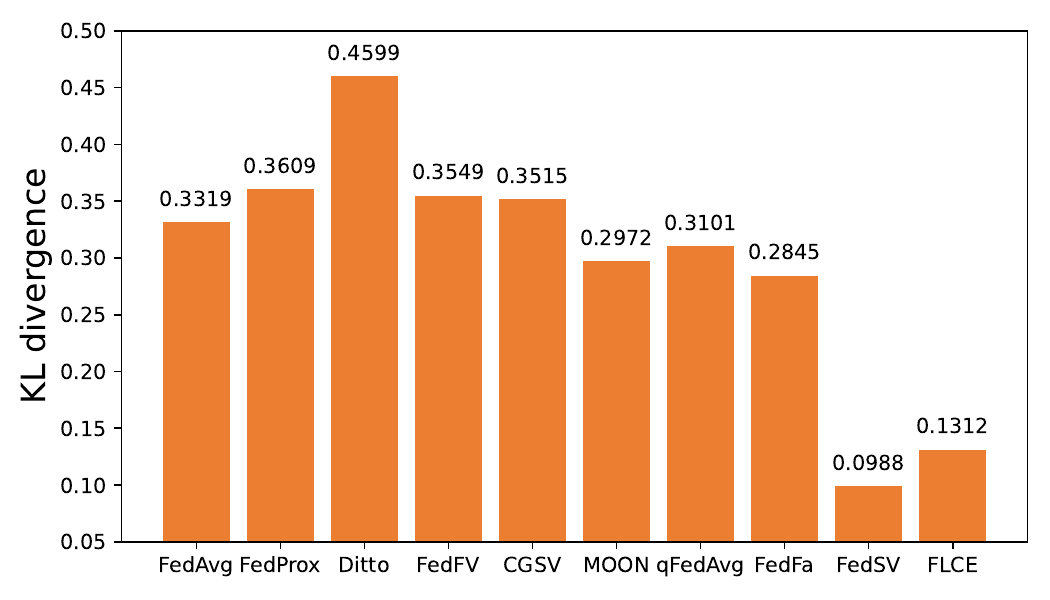}
	\caption{KL divergence between contribution evaluation results and data quality based on class diversity.}
	\label{fig.classes_weight_KL}
\end{figure}

\begin{figure}[t]
	\centering
	\includegraphics[width=0.9\linewidth]{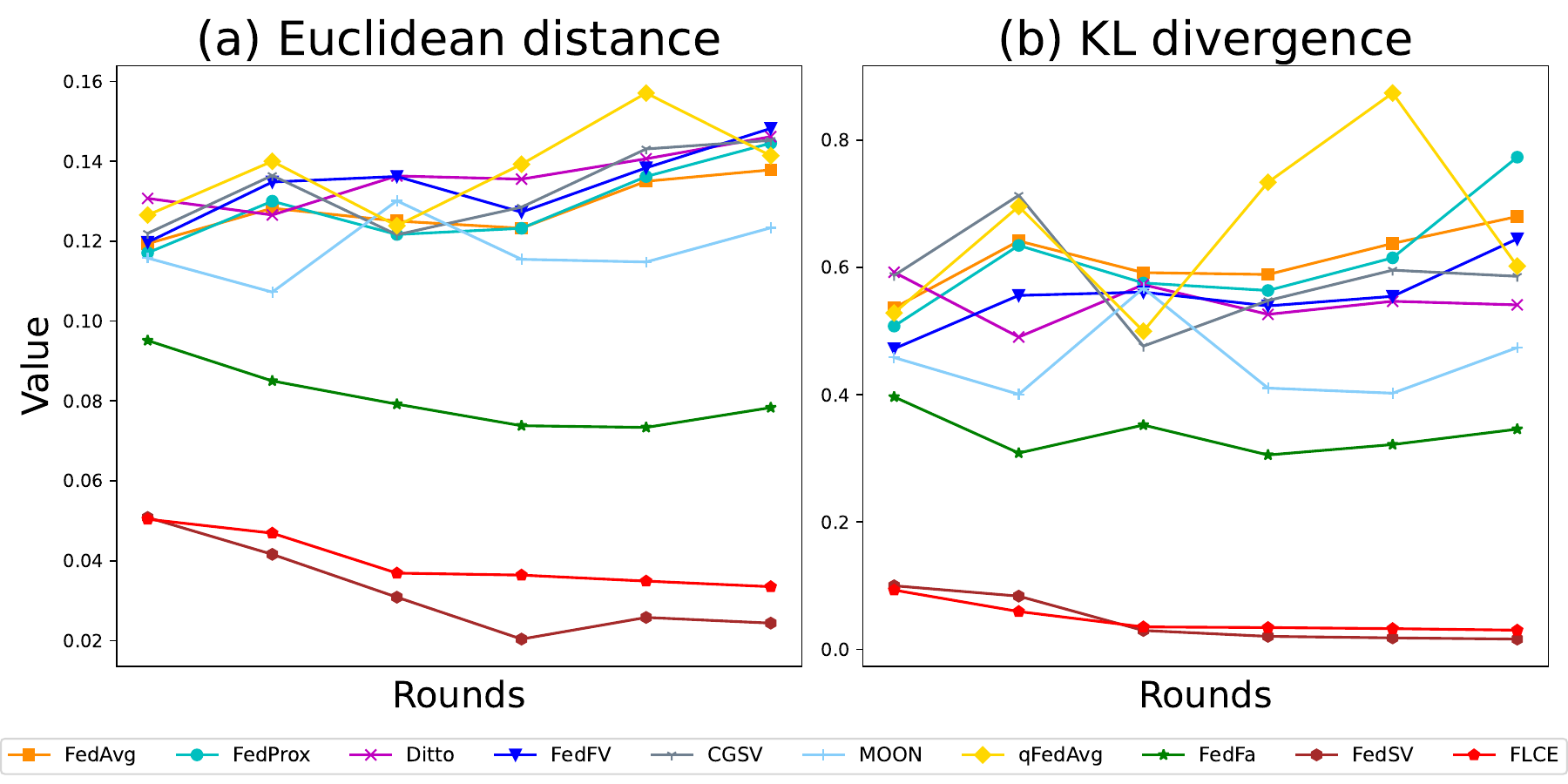}
	\caption{Differences between participant contributions calculated by each algorithm and canonical Shapley Value.}
	\label{Fig.distance_sv}
\end{figure}

\subsection{Effectiveness from Different Perspectives}
To further demonstrate the effectiveness of FLCE, we conducted a comprehensive evaluation from two additional perspectives: data quality based on class diversity and canonical Shapley value~\cite{shapley1953value} on the CIFAR-10 dataset with the Non-IID setting.

Firstly, existing research suggests that the quality of participants' data may be related to class diversity~\cite{xu2021validation}. Therefore, we incorporated the consideration of class diversity for each participant's data quality. Specifically, when calculating the diversity-based data quality for each participant, we multiplied the data volume by the ratio of the number of classes owned by the participant to the total number of classes. The results for all participants were then normalized. As illustrated in \figurename~\ref{fig.classes_weight_KL}, even when accounting for data class diversity, FLCE demonstrates the second-best performance, surpassed only by the FedSV method based on Shapley value calculations.

Secondly, the canonical Shapley value method is a classical approach in cooperative game theory for determining participants' contributions. Despite its computational intensity, its rationality in evaluating participant contributions is widely acknowledged. Therefore, we used the contribution evaluation results calculated by the canonical Shapley value method as a reference. Considering that some existing evaluation works employ Euclidean distance as a metric when using Shapley values~\cite{liu2022contribution}, we utilized both KL divergence and Euclidean distance in our assessment of different contribution evaluation methods. The experimental results are presented in \figurename~\ref{Fig.distance_sv}. \figurename~\ref{Fig.distance_sv}(a) and \figurename~\ref{Fig.distance_sv}(b) respectively show the changes in Euclidean distance and KL divergence between the contribution evaluation results computed by different methods and those of the canonical Shapley value method as the number of training rounds increases. The experimental results indicate that FLCE approaches the performance of Monte Carlo sampling-based FedSV, but with significantly reduced computational time (details in Section \ref{title.exp.Computational}).

These experimental findings from two distinct perspectives further validate the effectiveness of our proposed FLCE method in contribution evaluations.

\begin{figure}[t]
  \centering
  \subfloat[Global class level]
  {\includegraphics[width=0.24\textwidth]{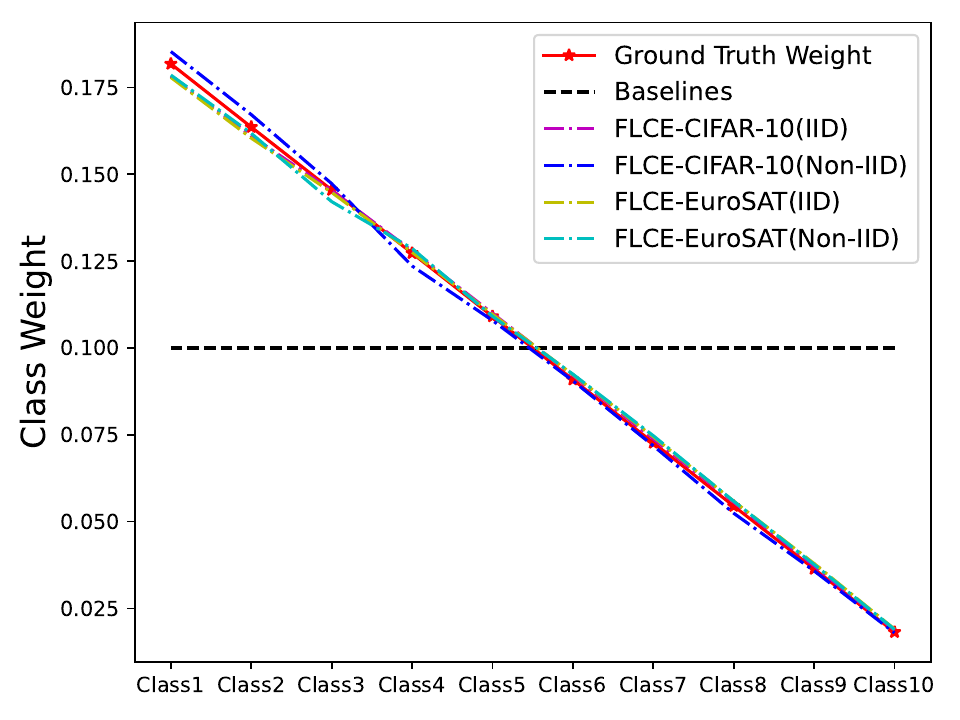}}    
  \subfloat[Local class level]
  {\includegraphics[width=0.24\textwidth]{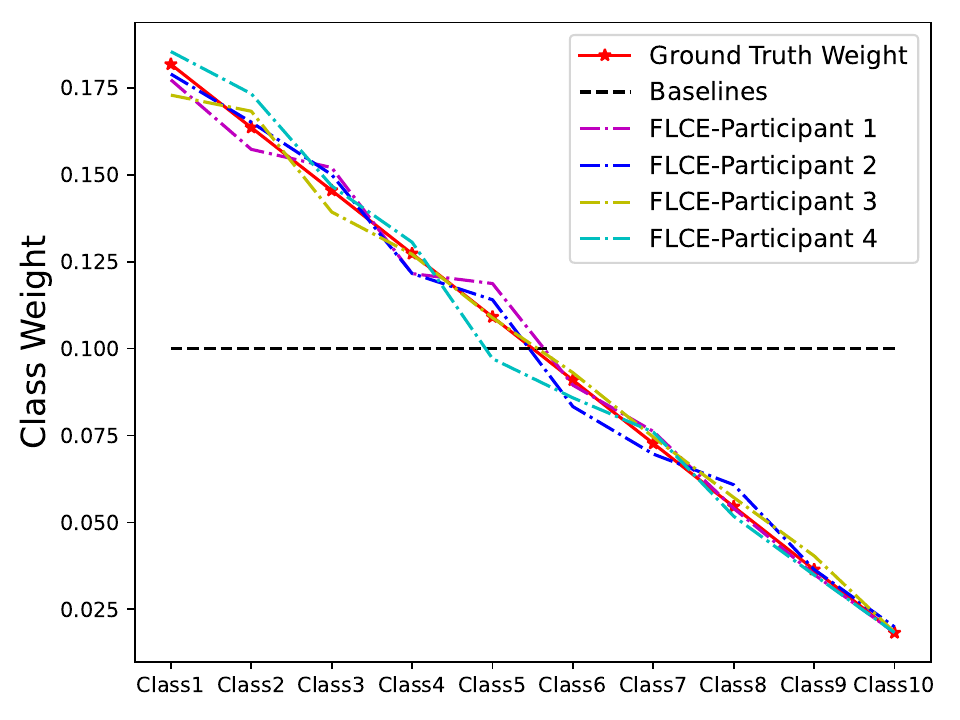}}
  \caption{Comparison of class contribution weights obtained by different methods and ground truth weight.}
  \label{Fig.class_weight}
\end{figure}
\subsection{Contribution Evaluation of Class Perspective}

Previous studies typically focused on evaluating contributions at the participant level, neglecting the class level. In the real world, the global model often prioritizes different classes variably, leading to disparities in class weights. FLCE excels not only at the participant level but also in evaluating contributions at the class level.

To further analyze contributions at the global and local class levels, we predefined individual weights for all classes, referred to as \emph{ground truth weight}. We then compared the class contribution weights determined by various methods with the ground truth weight on the CIFAR-10 and EuroSAT datasets, as shown in \figurename~\ref{Fig.class_weight}. The red solid line represents the ground truth weight. All baseline methods yield a uniform contribution weight of 0.1 for different classes, indicated by the green dashed line, because they overlook the weight variations among different classes. The contribution weight results for different classes obtained by FLCE in various scenarios are shown by the dotted and dashed lines.

As illustrated in \figurename~\ref{Fig.class_weight}(a), despite dataset and data distribution changes, FLCE's approach to evaluating weighted contributions of various classes globally remains highly effective. Furthermore, when focusing on individual class contributions within participants as depicted in \figurename~\ref{Fig.class_weight}(b), FLCE effectively evaluates class contributions with varying weights, delving into the details of individual participants. 
Contribution evaluation analysis from the global class level and local class level enhances the interpretability of the FLCE method. In contrast to earlier methods, FLCE can flexibly adjust class weights to effectively evaluate contributions in real-world scenarios.

\begin{figure}[t]
	\centering  
	\subfloat{
		\label{Fig.sub.1}
		\includegraphics[width=0.24\textwidth, height=0.23\textwidth]{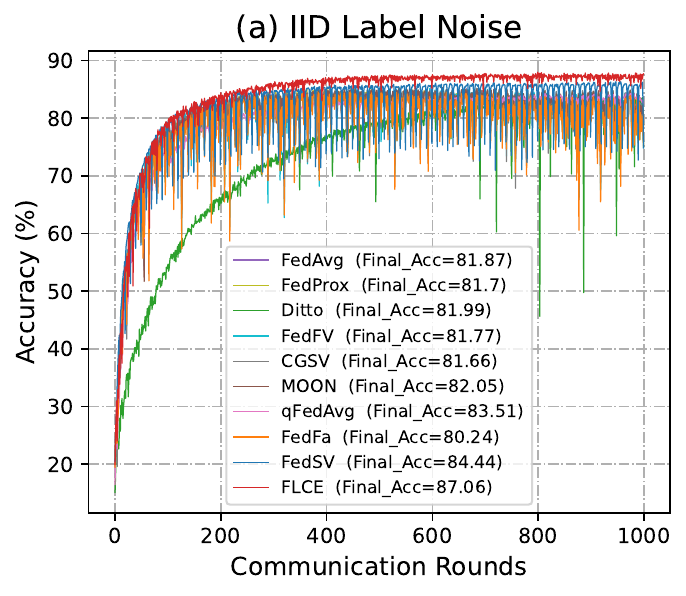}}
	\subfloat{
		\label{Fig.sub.2}
		\includegraphics[width=0.24\textwidth, height=0.23\textwidth]{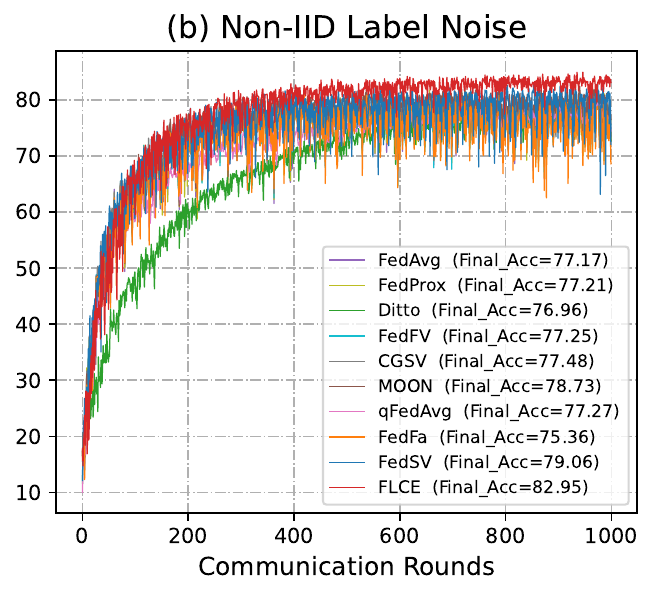}}
    \\
	\subfloat{
		\label{Fig.sub.3}
		\includegraphics[width=0.24\textwidth, height=0.23\textwidth]{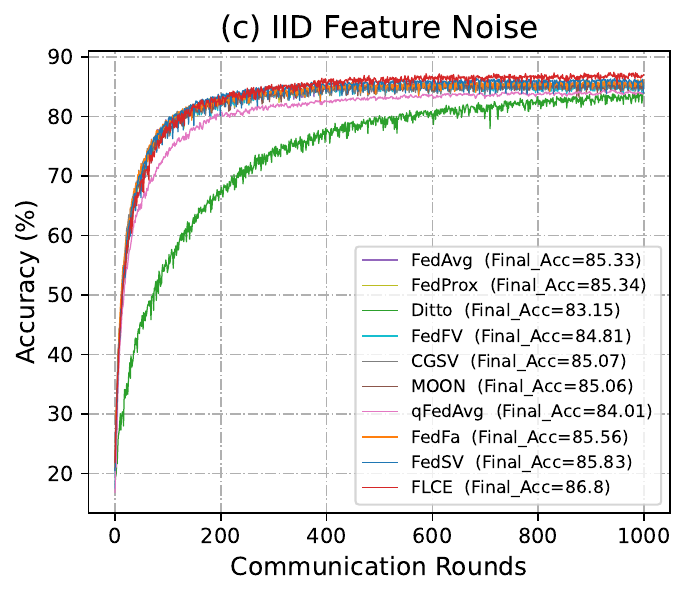}}
	\subfloat{
		\label{Fig.sub.4}
		\includegraphics[width=0.24\textwidth, height=0.23\textwidth]{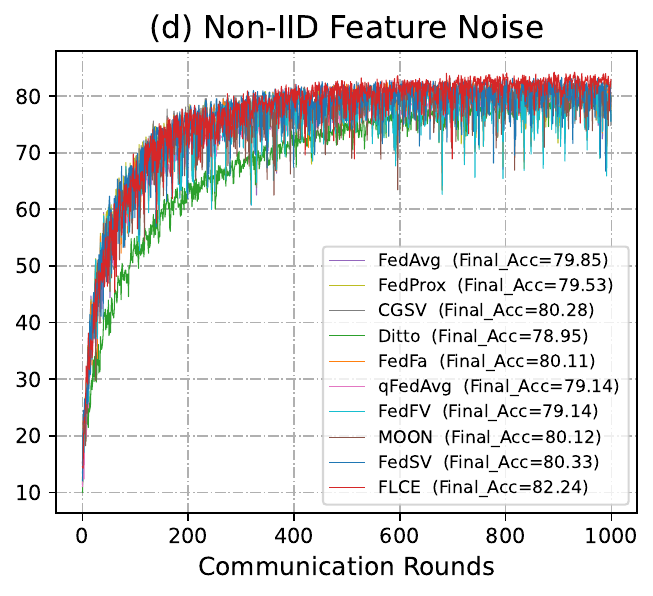}}
    \\
	\subfloat{
		\label{Fig.sub.5}
		\includegraphics[width=0.24\textwidth, height=0.23\textwidth]{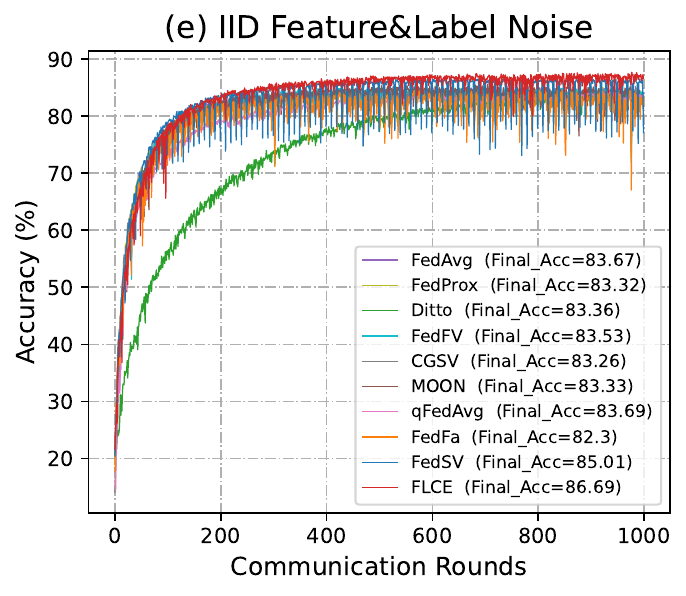}}
	\subfloat{
		\label{Fig.sub.6}
		\includegraphics[width=0.24\textwidth, height=0.23\textwidth]{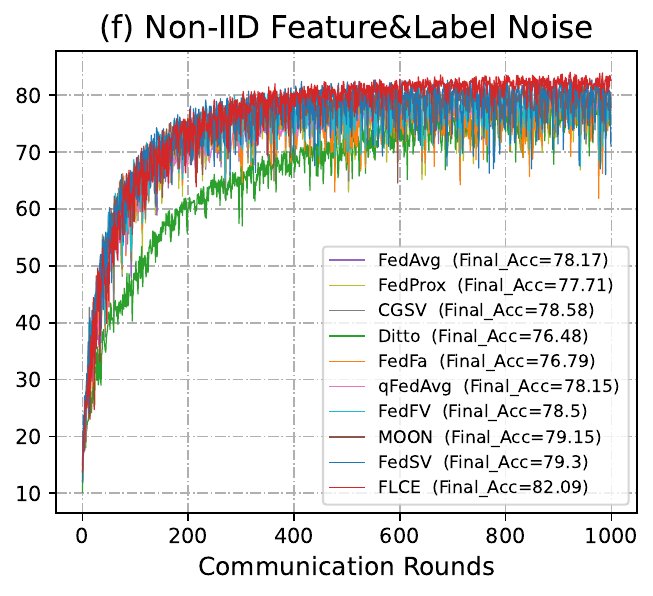}}
    \\

\caption{Accuracy of various methods on noisy datasets. Note: Final\_Acc represents the average accuracy over the final 100 rounds.}
	\label{fig.Noise}
\end{figure}

\begin{figure}[t]
  \centering
    \subfloat{
		\label{Fig.KL_IID_Noise}
		\includegraphics[width=0.24\textwidth]{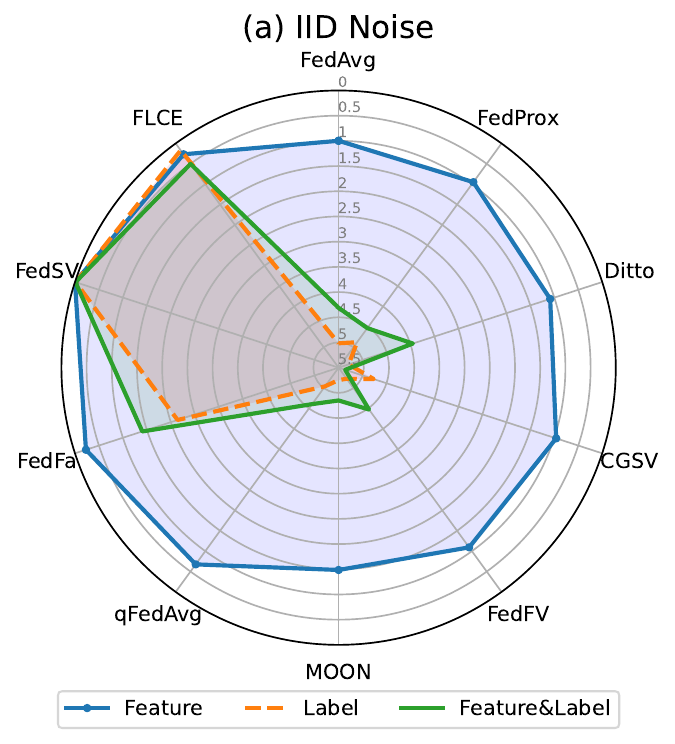}}
	\subfloat{
		\label{Fig.KL_NonIID_Noise}
		\includegraphics[width=0.24\textwidth]{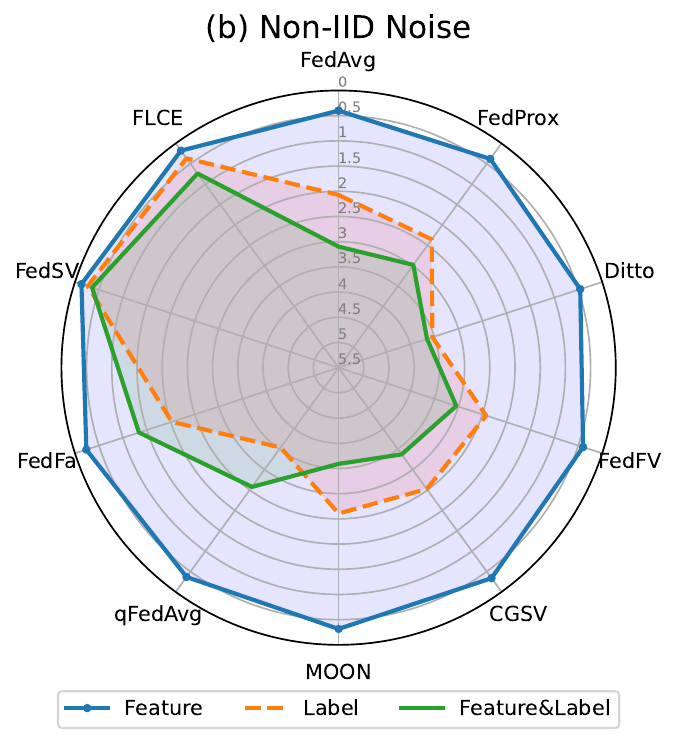}}
  \caption{KL divergence of various methods on noise datasets.}
  \label{fig.KL_Noise}
\end{figure}
\subsection{Contribution Evaluation of Noise Perspective}
In actual datasets, the presence of varying degrees of noise is also a significant form of heterogeneity. To assess FLCE's capability in handling noisy data, we simulated the scenarios by artificially injecting noise into features and labels. We categorized the noisy data into three types: feature noisy dataset, label noisy dataset, and dataset containing both noisy features and labels. We evaluated the accuracy and KL divergence of these datasets under IID and Non-IID conditions for CIFAR-10, as shown in \figurename~\ref{fig.Noise} and ~\ref{fig.KL_Noise}.

As illustrated in \figurename~\ref{fig.Noise}, the left column represents the accuracy of FL algorithms in IID scenarios, while the right column represents the accuracy in Non-IID scenarios. The KL divergence in various noise scenarios is shown in \figurename~\ref{fig.KL_Noise}. We can intuitively observe that FLCE achieves higher accuracy than other algorithms and exhibits smaller fluctuations during training. Meanwhile, FLCE also exhibits advantages in terms of KL divergence compared to other algorithms under noise scenarios. This robustness advantage is partly due to the construction and application of class contribution momentum. Specifically, by grouping prototypes of the same class and distancing those of different classes using contrastive learning, FLCE is less affected by noise compared to methods that rely on cross-entropy. Additionally, the server enhances robustness against low-quality participants by conducting a comprehensive evaluation of participants' contributions by category. Notably, FLCE's performance under label noise proves more effective than under feature noise, which supports the assertion about its strengths. 

\begin{table}[t]
\centering
\caption{Communication cost of FLCE.}
\begin{tabular}{c|ccc}
\hline
Dataset  & Prototype & Model  & Ratio  \\ \hline
CIFAR-10  & 640       & 272474 & 0.001174 \\ 
CIFAR-100 & 6400      & 278324 & 0.01136  \\ 
EuroSAT  & 640       & 272474 & 0.001174 \\ \hline
\end{tabular}
\label{tab: Communication Cost}
\end{table}

The experimental results indicate that FLCE exhibits excellent performance in noisy scenarios due to its emphasis on the intrinsic properties of the data and its reduced susceptibility to errors in labeling. This reflects the fundamental rationality and effectiveness of FLCE in handling noisy heterogeneous scenarios.

\subsection{Communication Cost}

During the training process, FLCE requires participants to compute prototypes for each class, which means participants need to upload not only their models but also prototypes for each class. 

However, the contribution of prototypes to overall communication in each round is very small. For instance, taking FLCE using ResNet20 on the CIFAR-10, CIFAR-100, and EuroSAT. For instance, the model contains 272474 parameters to upload and download on CIFAR-10, while the size of each prototype is 64 with a total of 10 classes, resulting in a total prototype size of 640, accounting for only 0.12\% of the total communication in each round as \tablename~\ref{tab: Communication Cost}. We can directly observe that the additional upload of prototypes by participants only accounts for a very small portion of the total communication cost.

Therefore, the communication cost incurred by uploading prototypes is minimal for participants, but it can significantly improve fidelity and effectiveness.

\subsection{Computational Cost}
\label{title.exp.Computational}
\begin{figure}[t]
	\centering
	\includegraphics[width=0.9\linewidth]{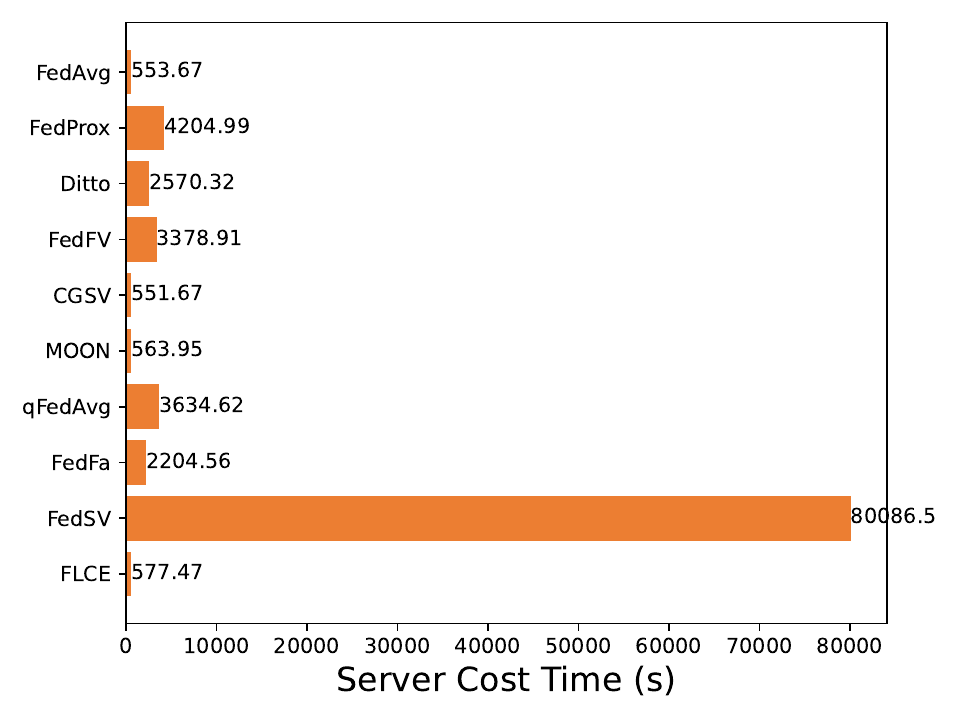}
	\caption{Computational cost of different methods.}
	\label{fig.computational_cost}
\end{figure}

Typically, the server in FL has higher performance capabilities than participants' local devices, making additional computations on the server a viable strategy to enhance model performance. During the global update process, computational cost varies depending on the algorithm used. We conducted tests to measure the time required by the server on the CIFAR-10 dataset with the Non-IID setting, as shown in \figurename~\ref{fig.computational_cost}.
On the one hand, FedAvg requires simple aggregation at the server, resulting in minimal time consumption. On the other hand, FedSV requires significant time for computing the Shapley values at the server, which involves arranging and combining participant models. As shown in \figurename~\ref{fig.computational_cost}, FLCE's time consumption is comparable to FedAvg, demonstrating its efficiency in computation. Because the prototypes are extracted from the participants' data, the smaller size of the prototypes results in minimal computational overhead. 
The experimental results indicate that FLCE exhibits a time-cost advantage and is an efficient contribution evaluation method for heterogeneous participants in FL.

\subsection{The Impact of Statistical Heterogeneity}
\begin{figure}[t]
	\centering
	\includegraphics[width=0.95\linewidth]{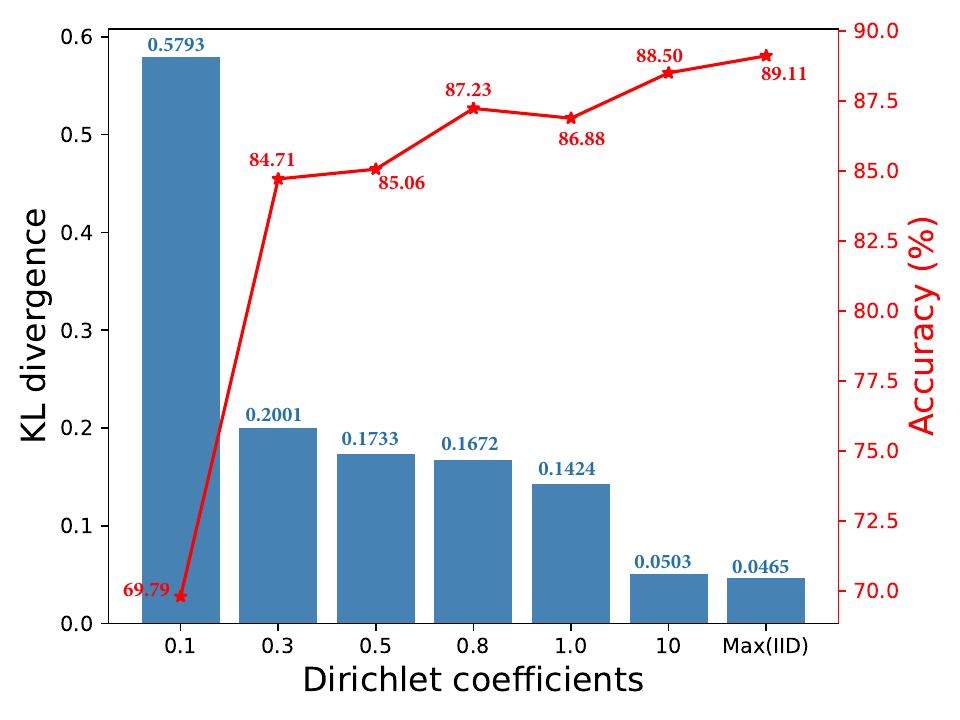}
	\caption{The impact of different heterogeneous scenarios.}
	\label{fig:heterogeneous}
\end{figure}

The statistical heterogeneity of participant data in the real world can significantly impact algorithm performance. To assess the effect of statistical heterogeneity on FLCE, we used the CIFAR-10 dataset with varying Dirichlet coefficients $\delta$ to measure FLCE's performance, as shown in \figurename  ~\ref{fig:heterogeneous}.

Our observations reveal that as data statistical heterogeneity varies, both accuracy and KL divergence systematically change. When statistical heterogeneity is at its maximum ($\delta$=0.1), participants exhibit the lowest accuracy 69.79\%, and the highest KL divergence 0.5793. As $\delta$ increases, indicating reduced heterogeneity, both accuracy and KL divergence improve, with accuracy peaking at 89.11\% and KL divergence minimizing to 0.0465 when $\delta=Max$.
The experimental results indicate that FLCE's performance gradually improves as the statistical heterogeneity decreases, demonstrating a consistent pattern when facing data of varying degrees of heterogeneity. This highlights FLCE's robust performance across a spectrum of statistical heterogeneity, confirming its effectiveness in diverse federated environments.

\section{Conclusion}
In this work, we propose the first contribution evaluation method via participants' representations and introduce a novel contribution evaluation indicator class contribution momentum. We adopt a tripartite perspective to conduct contribution evaluation, encompassing the individual, relative, and overall perspectives. The server can effectively and efficiently evaluate participants' contributions by leveraging representations extracted from their heterogeneous data. The results of numerous experiments demonstrate that FLCE performs excellently in various heterogeneous scenarios. Moreover, as far as we know, we are the first to achieve contribution evaluation at the class level, which is a common real-world scenario. In addition, due to our FLCE adopting an original FL framework, participants only need to compute representations of their local data, making it versatile and scalable.



\bibliographystyle{ACM-Reference-Format}
\bibliography{sample}

\end{document}